\documentclass[letterpaper, 10 pt, journal, twoside]{IEEEtran}  

\IEEEoverridecommandlockouts                              

\usepackage{lipsum}
\usepackage{mathtools}
\usepackage{cuted}
\usepackage[utf8]{inputenc}
\usepackage[english]{babel}
\usepackage{wrapfig}

\newtheorem{proposition}{Proposition}
\usepackage{xcolor}
\usepackage{enumitem}
\usepackage{amssymb}
\usepackage{amsmath} 

\newcommand{\SB}[1]{\textcolor{black}{#1}}

\usepackage{graphicx} 
\usepackage{hyperref}
\usepackage{multirow}

\usepackage{multicol}

\title{
Reinforcement Learning for POMDP: Partitioned Rollout and Policy Iteration with Application to Autonomous Sequential Repair Problems
}
\author{Sushmita Bhattacharya$^{1}$, Sahil Badyal$^{1}$, Thomas Wheeler$^{1}$, Stephanie Gil$^{1,2}$, Dimitri Bertsekas$^{3,4}$
\thanks{Manuscript received: September, 10, 2019; Revised December, 10, 2019;
Accepted January 23, Day, 2020.}
\thanks{The  authors  gratefully  acknowledge  support
by  the  National  Science  Foundation  CAREER  Award  [Grant  No.
1845225]}
\thanks{$^{1}$REACT Lab Arizona State University, 699 S Mill Avenue, Tempe, AZ 
{\tt\small sbhatt55@asu.edu}}%
\thanks{$^{2}$Assistant Professor of Computer Science, Arizona State University, 699 S Mill Avenue, Tempe AZ 85287
{\tt\small sgil@asu.edu}}%
\thanks{$^{3}$ McAfee Professor of Engineering, Massachusetts Institute of Technology, 77 Mass Ave Cambridge, MA 02139
{\tt\small dimitrib@mit.edu}}%
\thanks{$^{4}$ Fulton Professor of Computational Decision Making, Arizona State University, 699 S Mill Avenue, Tempe AZ 85287
{\tt\small dbertsek@asu.edu}}%
}

\begin{document}
\twocolumn[
\Large\textbf{IEEE Copyright Notice}\\\\
 \textcopyright 2020 IEEE.  Personal use of this material is permitted.  Permission from IEEE must be obtained for all other uses, in any current or future media, including reprinting/republishing this material for advertising or promotional purposes, creating new collective works, for resale or redistribution to servers or lists, or reuse of any copyrighted component of this work in other works.\\\\
 Accepted to be published in: IEEE Robotics and Automation Letters 2020.
 ]
\maketitle
\thispagestyle{plain}
\pagestyle{plain}

\begin{abstract}
In this paper we consider infinite horizon discounted dynamic programming problems with finite state and control spaces, and partial state observations. We discuss an algorithm that uses multistep lookahead, truncated rollout with a known base policy, and a terminal cost function approximation. This algorithm is also used for policy improvement in an approximate policy iteration scheme, where successive policies are approximated by using a neural network classifier. A novel feature of our approach is that it is well suited for distributed computation through an extended belief space formulation and the use of a partitioned architecture, which is trained with multiple neural networks. We apply our methods in simulation to a class of sequential repair problems where a robot inspects and repairs a pipeline with potentially several rupture sites under partial information about the state of the pipeline.
\end{abstract}


\def\a{\alpha}
\def\be{\beta}
\def\b{\beta}
\def\l{\lambda}
\def\g{\gamma}
\def\m{\mu}
\def\p{\pi}
\def\r{\rho}
\def\e{\epsilon}
\def\t{\tau}
\def\d{\delta}
\def\s{\sigma}
\def\f{\phi}
\def\tl{\tilde}
\section{Introduction}
\IEEEPARstart{W}{e} consider the classical partial observation Markovian decision problem (POMDP) with a finite number of states and controls, and discounted additive cost over an infinite horizon. The optimal solution is typically intractable, and several suboptimal solution/reinforcement learning approaches have been proposed. Amongst these, are point-based value iteration (see e.g., \cite{PGT06, SpV05,SPK13}), approximate policy iteration methods based on the use of finite state controllers (see e.g., \cite{KLC98, Han98, MPK99}), the use of Monte Carlo tree search and adaptive sampling methods for multistep lookahead with terminal cost function approximation (see e.g., \cite{RPP08,SiV10}), and discretization/aggregation methods based on the solution of a related perfectly observable Markovian decision problem (see e.g., \cite{ZhH01, YuB04, Ber19}). There have also been proposals of policy gradient and related actor-critic methods (see \cite{BaB01, Yu05, ELP12}) that are largely unrelated to methodology proposed here.

In this paper we focus on methods of policy iteration (PI) that are based on rollout, and approximations in policy and value space. They update a policy by using truncated rollout with that policy, and a terminal cost function approximation. Several earlier works include elements of our algorithmic approach. In particular, related methods have been applied with some variations to perfect state information problems, notably in the backgammon algorithm of Tesauro and Galperin \cite{TeG96}, and in the AlphaGo program \cite{SHM16}, as well as in the classification-based PI algorithm of Lagoudakis and Parr \cite{LaP03} (see also \cite{DiL08, LGM10, LiW14}). The AlphaZero program also involves a similar approximation architecture, but in its published description \cite{SHS17}, it does not use rollout, likely because the use of long multistep lookahead in conjunction with Monte Carlo tree search makes the use of rollout unnecessary. A further novel feature of our algorithms is the use of a partitioned architecture, involving multiple policy and value neural networks, which is well-suited for distributed implementation. Partitioning in conjunction with asynchronous PI was originally proposed by Bertsekas and Yu \cite{BeY10}, and further developed in the papers \cite{BeY12} and \cite{YuB13} [see also the books \cite{Ber12} (Section 2.6) and \cite{Ber18} (Section 2.6) for descriptions, analysis, and extensions]. However, most of this research was focused on the case of perfect state information, and lookup table representations of policies and cost functions. Thus, while most of the principal elements of our approach have individually appeared in various forms in earlier perfect state information algorithmic frameworks, their combination has not been brought together into a single algorithm, and moreover they have not been adapted to the special challenges posed by POMDP.

Because of its simulation-based rollout character, the methodology of this paper depends critically on the finiteness of the control space, but it does not rely on the piecewise linear structure of the finite horizon optimal cost function of POMDP. It can be extended to POMDP with infinite state space but finite control space, although we have not considered this possibility in this paper. We describe error bounds to guide the implementation of our algorithms, and we provide results of computational experimentation showing that our methods are viable and well-suited to the POMDP structure. 

We apply our methods to a class of problems in robotics, involving sequential repairs, and search and rescue, where the POMDP model is particularly well-suited to deal with partial state information. Autonomous robots in search and rescue have been viewed as one of the robotics applications where POMDP approaches need further development and where these approaches can have great impact~\cite{SARuavs, Cassandra2003,decisionMakingAutonomousVeh, controlRusUnderwater, serviceRobotGaurav,decisionMakingRus}. 
Indeed, exploration and learning in unknown environments has been identified as one of the grand challenges of Science Robotics~\cite{grandChallengesWood}. These problems are very complex, characterized by large state spaces and constrained communication. Many tools in machine learning and artificial intelligence have recently been developed for tackling difficult problems in robotics~\cite{ AIHowDeepRL, RLWolfram, deepLearningAIAbeel, AIDeliberation}. Of particular relevance to the current paper are those works that involve decision making under uncertainty and POMDP frameworks~\cite{POMDPAmatoDecentralized, POMDPSycara}.

As an application of our methodology, we consider a robot that must decide on the sequence of linearly arranged pipeline locations to explore and/or repair using prior information and observations made \emph{in situ}. The actual damage of each location is initially unknown and can become worse if not repaired. This process is modeled by a Markov chain with known transition probabilities. The problem has a very large number of states (\SB{$\approx 10^{26}$ in our largest implementation}). Our experiments demonstrate that our methodology is well suited to robotics applications that involve: i) large state spaces and long planning horizons, which exacerbate both the curse of dimensionality and the curse of history, and ii) decaying environments as in sequential repair problems, where it is important to use a policy that 
can identify and execute critical actions in minimum time. 

We compare the performance of our proposed method with the POMCP and DESPOT methods from~\cite{SiV10, despot}, respectively, and we showcase the generality of our method by applying it to more complex versions of the pipeline problem: a two-dimensional grid pipeline, and a multi-robot variant of the problem. The use of distributed computation within our framework also suggests future applicability to multi-robot problems with asynchronous communication and/or bandwidth constraints, which are of great importance in robotics~\cite{gilMultiRobot,gilISRR2019, gilMultiRobotCov, controlRusUnderwater, POMDPAmatoDecentralized}.

We also provide theoretical support for our methodology in the form of a performance bound (Prop.\ 1), which shows improvement of the rollout policy over the base policy (approximately). Alternative methodologies, such as POMCP and DESPOT, do not enjoy a comparable level of theoretical support.

In summary, the contributions of the paper include: 1) the development of an algorithmic framework for finding approximately optimal policies for large state space POMDP, including the development of distributed PI methods with a partitioned architecture, 2) performance bounds to support the architectural structure, and 3) implementation and validation in simulation of the proposed methods on a pipeline repair problem, whose character is shared by broad classes of problems in robotics.

\vspace{-0.04in}
\section{Extended Belief Space Problem Formulation}
The starting point of our paper is the classical belief space formulation of a POMDP. However, we extend this formulation to make it compatible with a distributed multiprocessor implementation of our algorithm. In particular, we propose to use as state a sufficient statistic that subsumes the belief state in the sense that its value determines the value of the belief state. In this section, we describe this extended belief space problem formulation.

We assume that there are $n$ states denoted by $i\in\{1,\ldots,n\}$ and a finite set of controls $U$ at each state. We denote by $p_{ij}(u)$ the transition probability from $i$ to $j$ under $u\in U$, and by $g(i,u,j)$ the corresponding transition cost. The cost is accumulated over an infinite horizon and is discounted by $\alpha\in(0,1)$. At each new generated state $j$, an observation $z$ from a finite set $Z$ is obtained with known probability $p(z\mid j,u)$ that depends on $j$ and the control $u$ that was applied prior to the generation of $j$. The objective is to select each control optimally as a function of the prior history of observations and controls.

A classical approach to this problem is to convert it to a perfect state information problem whose state is the current belief $b=\big(b(1),\ldots,b(n)\big)$, where $b(i)$ is the conditional distribution of the state $i$ given the prior history. It is well-known that $b$ is a sufficient statistic, which can serve as a substitute for the set of available observations, in the sense that optimal control can be achieved with knowledge of just $b$. In this paper, we consider a more general form of sufficient statistic, which we call the {\it feature state} and we denote by $y$. {\it We require that the feature state $y$ subsumes the belief state $b$.} By this we mean that $b$ can be computed exactly knowing $y$. One possibility is for $y$ to be the union of $b$ and some identifiable characteristics of the belief state, or some compact representation of the measurement history up to the current time (such as a number of most recent measurements, or the state of a finite-state controller). 
We also make the additional assumption that {\it $y$ can be sequentially generated using a known feature estimator $F(y,u,z)$\/}. By this we mean that given that the current feature state is $y$, control $u$ is applied, and observation $z$ is obtained, the next feature can be exactly predicted as $F(y,u,z)$.

Clearly, since $b$ is a sufficient statistic, the same is true for $y$. 
Thus the optimal costs achievable by the policies that depend on $y$ and on $b$ are the same. However, specific suboptimal schemes may become more effective with the use of the feature state $y$ instead of just the belief state $b$. Moreover, the use of $y$ can facilitate the use of distributed computation through partitioning of the space of features $y$, as we will explain later.

The optimal cost $J^*(y)$, as a function of the sufficient statistic/feature state $y$, is the unique solution of the corresponding Bellman equation $J^*(y)=(TJ^*)(y)$ for all $y$, where $T$ is the Bellman operator 
\vspace{-0.05in}
$$(TJ)(y)=\min_{\mu\in {\cal M}}(T_{\mu} J)(y),$$
with ${\cal M}$ being the set of all stationary policies [functions $\m$ that map $y$ to a control $\m(y)\in U$], and $T_{\mu}$ being the Bellman operator corresponding to $\mu$:
\vspace{-0.05in}
$$(T_\mu J)(y)=\hat g(y,\mu(y))+\alpha \sum_{z\in Z}\hat p(z\,|\, b_y,\mu(y)) J\big(F(y,\mu(y),z)\big).$$
Here $b_y$ denotes the belief state that corresponds to feature state $y$, $\hat g(y,u)$ is the expected cost per stage
$$\hat g(y,u)=\sum_{i=1}^n b_y(i)\sum_{j=1}^n p_{ij}(u)g(i,u,j),$$
$\hat p(z\,|\, b_y,u)$ is the conditional probability that the next observation will be $z$ given the current belief state $b_y$ and control $u$, and $F$ is the feature state estimator.

The feature space reformulation of the problem can serve as the basis for approximation in value space, whereby $J^*$ is replaced in Bellman's equation by some function $\tilde J$ after one-step or multistep lookahead. For example a one-step lookahead scheme yields the suboptimal policy $\tilde \mu$ given by
\vspace{-0.05in}
$$\tilde \mu(y)\in\arg\min_{u\in U}\left[\hat g(y,u)+\alpha \sum_{z\in Z}\hat p(z\,|\, b_y,u) \tilde J\big(F(y,u,z)\big)\right].\eqno(1)$$
In $\ell$-step lookahead schemes, $\tilde J$ is used as terminal cost function in an $\ell$-step horizon version of the original infinite horizon problem (see e.g., \cite{Ber19}). In the standard form of a rollout algorithm, $\tilde J$ is the cost function of some base policy. Here we adopt a rollout scheme with $\ell$-step lookahead, which involves rollout truncation and terminal cost approximation. This scheme has been formalized and discussed in the book \cite{Ber19} (Section 5.1), and is described in the next section.

\vspace{-0.04in}
\section{Truncated Rollout with Terminal Cost Function Approximation}
\begin{figure}[ht]
    \centering
    \includegraphics[width=0.48\textwidth]{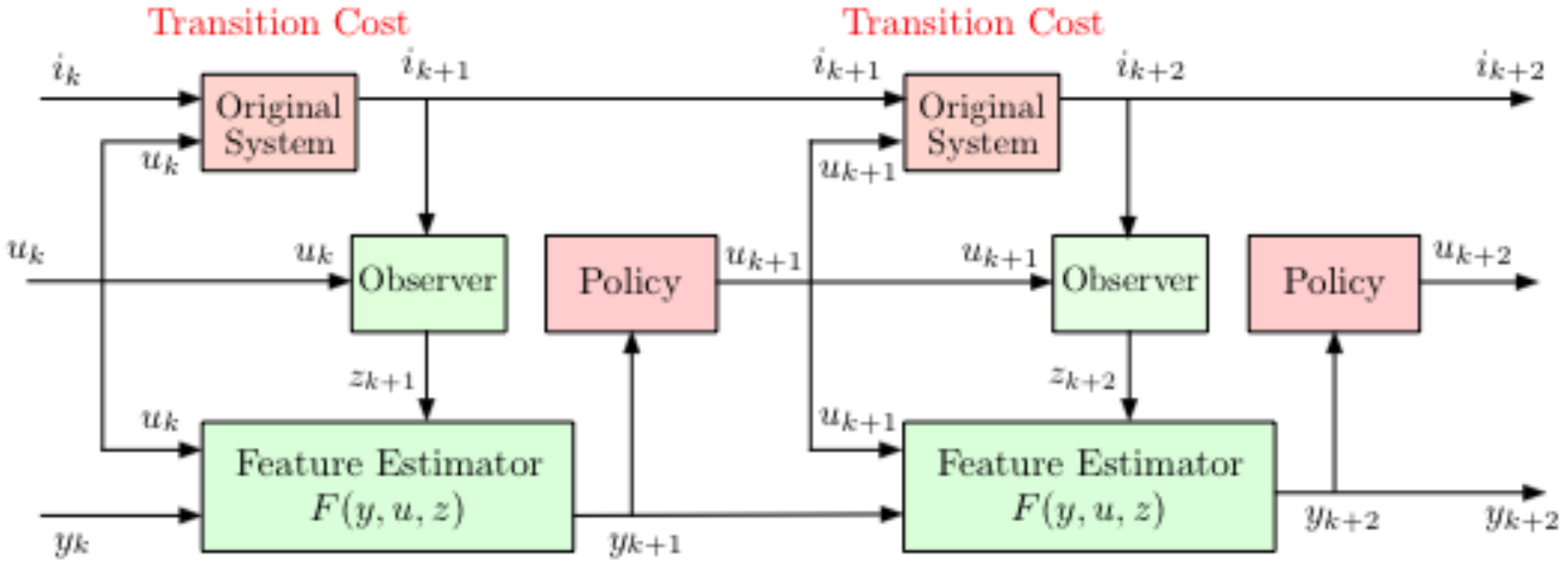}
    \caption{\vspace{-0.0in} Composite system simulator for POMDP for a given policy. The starting state $i_k$ at stage $k$ of a trajectory is generated randomly using the belief state $b_k$, which is in turn computed from the feature state $y_k$.}
    \label{belief_simulator}
\end{figure}

\begin{figure}

\vspace{0.05in}
\centering
\includegraphics[scale=0.35]{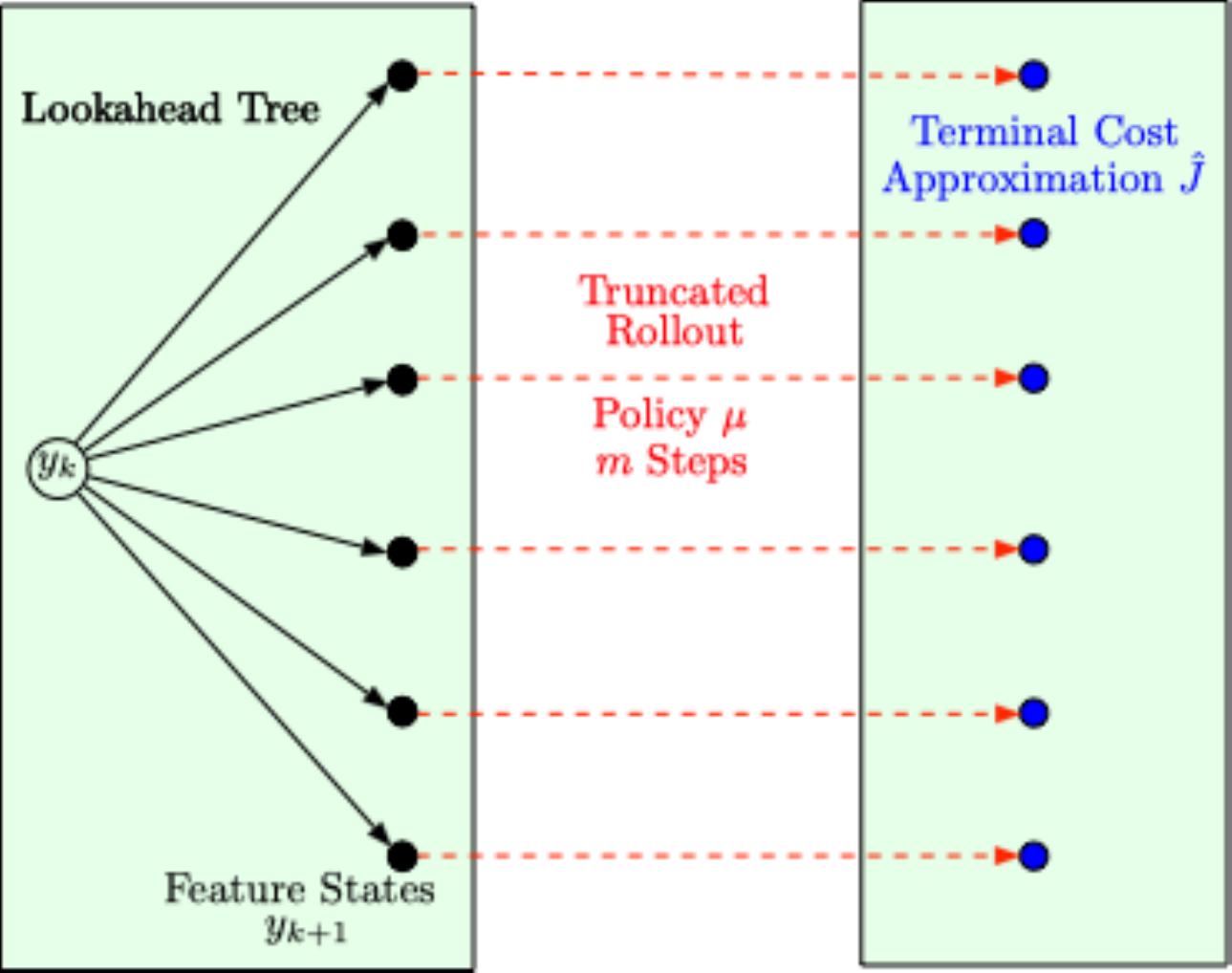}
    \caption{A truncated rollout scheme: One-step lookahead is followed by $m$-step rollout with base  policy $\mu$, and terminal cost approximation $\hat{J}$.}
    \label{fig:lookahead}
\end{figure}

In the pure form of the rollout algorithm, the cost function approximation $\tilde{J}$ is the cost function $J_\mu$ of a known policy $\mu$, called the {\it base policy\/}, and its value $\tilde J(y)=J_\mu(y)$ at any $y$ is obtained by first extracting $b$ from $y$, and then running a simulator starting from $b$, and using the system model, the feature generator, and $\mu$ (see Fig. \ref{belief_simulator}). In the truncated form of rollout, $\tilde J(y)$ is obtained by running the simulator of $\mu$ for a given number of steps $m$, and then adding a terminal cost approximation $\hat J(\bar y)$ for each terminal feature state $\bar y$ that is obtained at the end of the $m$ steps of the simulation with $\mu$ (see Fig. \ref{fig:lookahead} for the case where $\ell=1$). 



Thus the rollout policy is defined by the base policy $\mu$, the terminal cost function approximation $\hat J$, the number of steps $m$ after which the simulated trajectory with $\mu$ is truncated, as well as the lookahead size $\ell$; see Fig. \ref{fig:lookahead}. The choices of $m$ and $\ell$ are typically made by trial and error, based on computational tractability among other considerations, while $\hat J$ may be chosen on the basis of problem-dependent insight or through the use of some off-line approximation method. In variants of the method, the multistep lookahead may be implemented approximately using a Monte Carlo tree search or adaptive sampling scheme. In our experiments, we used a variable and state-dependent value of $m$. 
Monte-Carlo tree search did not work well in our initial experiments, and we subsequently abandoned it. 

Using $m$-step rollout between the $\ell$-step lookahead and the terminal cost approximation gives the method the character of a single PI. We have used repeated truncated rollout as the basis for constructing a PI algorithm, which we will discuss shortly (see also \cite{Ber19}, Section 5.1.2).

There are known error bounds that can be applied to the preceding truncated rollout scheme. These bounds were given in \cite{Ber19}, Section 5.1, Prop. 5.1.3, but they were derived for the case where the number of states is finite, so they do not apply to our feature state representation of a POMDP. However, the bounds can be extended to our infinite feature space case, since the proof arguments of \cite{Ber19} do not depend on the finiteness of the state space. We thus give the bounds without proof.

\proposition{(Error Bounds)} Consider a truncated rollout scheme consisting of $\ell$-step lookahead, followed by rollout with a policy $\mu$ for $m$ steps, and a terminal cost function approximation $\hat{J}$ at the end of $m$ steps. Let $\tilde{\mu}$ be the policy generated by this scheme. Then: 
{(a)} We have
\vspace{-0.07in}
$$\|J_{\tilde{\mu}}-J^*\|\leq \frac{2\alpha^\ell}{1-\alpha}\|T^m_\mu \hat{J}-J^*\|,$$
where $T^m_\mu \hat{J}$ is the result of applying $m$ times to $\hat{J}$ the Bellman operator $T_\mu$ for policy $\mu$, and $\|\cdot\|$ is the sup norm on the space of bounded functions of the feature state $y$.
{(b)} We have for all $y$
\vspace{-0.1in}
$$J_{\tilde{\mu}}(y)\leq J_\mu(y)+\frac{2}{1-\alpha}\|\hat{J}-J_\mu\|.$$
\vspace{-0.1in}

The first bound implies that as the size of lookahead $\ell$ increases, the bound on the performance of the rollout policy improves. 
The second bound suggests that if $\hat J$ is close to $J_\mu$, the performance of the rollout policy $\tilde \mu$ is approximately improved (to within an error bound), relative to the performance of the base policy $\mu$. This is typical of the practically observed cost improvement property of rollout schemes. In particular, when $\hat J=J_\mu$ we obtain $J_{\tilde{\mu}}\leq J_\mu$, which is the theoretical policy improvement property of rollout (see \cite{Ber19}, Section 5.1.2). 
\vspace{-0.04in}
\section{Supervised Learning of Rollout Policies and Cost Functions - Policy Iteration}
\noindent 
The rollout algorithm uses multistep lookahead and on-line simulation of the base policy to generate the rollout control at any feature state of interest. To avoid the cost of on-line simulation, we can approximate the rollout policy off-line by using some approximation architecture that may involve a neural network. This is policy approximation built on top of the rollout scheme (see \cite{Ber19}, Sections 2.1.5 and 5.7.2).

To this end, we introduce a parametric family/architecture of policies of the form $\hat
\mu(y,r)$, where $r$ is a parameter vector. We then construct a training set that consists of a large number of sample feature state-control pairs $(y^s,u^s)$, $s=1,\ldots,q$, such that for each $s$, $u^s$ is the rollout control at feature state $y^s$. We use this data set to obtain a parameter $\bar r$ by solving a corresponding classification problem, which associates each feature state $y$ with a control $\hat
\mu(y,\bar r)$. The parameter $\bar r$ defines a classifier, which given a feature state $y$, classifies $y$ as requiring control $\hat \mu(y,\bar r)$ (this idea was proposed in the context of PI in the paper \cite{LaP03}, and is also described in the book \cite{Ber19}, Section 3.5).
The classification problem is often solved with the use of neural networks, and this has been our approach in our experimentation.
\vspace{-0.04in}
\subsection{Approximate Policy Iteration (API)}
\begin{figure}[h]
    \centering
    
    \includegraphics[width=1\linewidth, height=4cm]{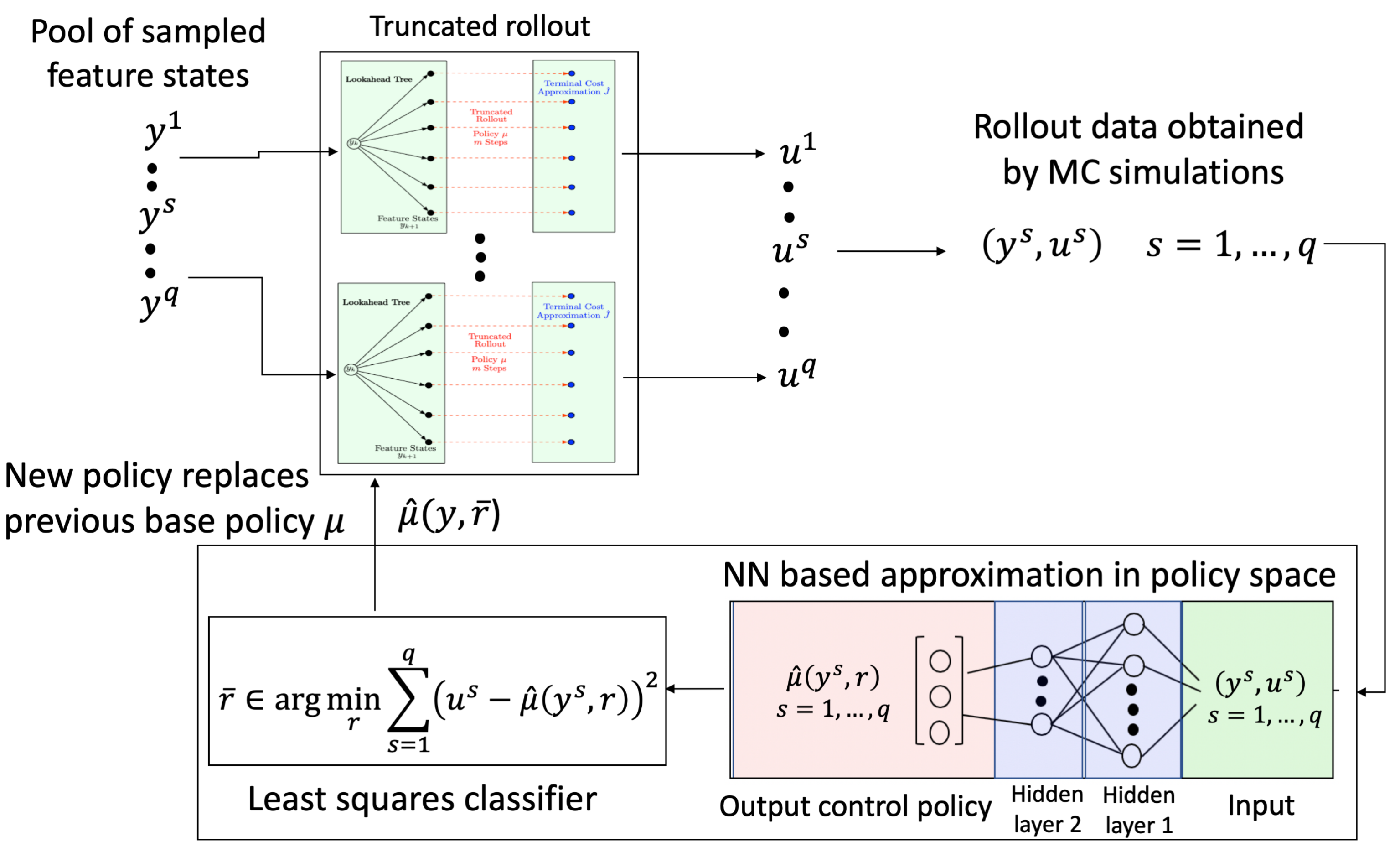} 
    \caption{\vspace{-0.01in} Approximate PI scheme based on rollout and approximation in policy space. 
    }
    \label{fig:PI}
\end{figure}

\noindent
We can also apply the rollout policy approximation to the context of PI. The idea is to view rollout as a one-step policy improvement, and to view the PI algorithm as a {\it perpetual rollout process\/}, which performs multiple policy improvements, using at each iteration the current policy as the base policy, and the next policy as the corresponding rollout policy. In particular, we consider a PI algorithm where at the typical iteration we have a policy $\mu$, which we use as the base policy to generate many feature state-control sample pairs $(y^s,u^s)$, $s=1,\ldots,q$, where $u^s$ is the rollout control corresponding to \SB{feature state $y^s$}. We then obtain an ``improved" policy $\hat \mu(y,\overline r)$ with an approximation architecture and a classification algorithm, as described above. The ``improved" policy is then used as a base policy to generate samples of the corresponding rollout policy, which is approximated in policy space, etc; see Fig. \ref{fig:PI}.

To use truncated rollout in this PI scheme, we must also provide a terminal cost approximation, which may take a variety of forms. Using zero is a simple possibility, which may work well if either the size $\ell$ of multistep lookahead or the length $m$ of the rollout is relatively large. Another possibility, employed in our pipeline repair problem, is to use as terminal cost in the truncated rollout an approximation of the cost function of some base policy, which is obtained with a neural network-based approximation architecture.

In particular, at any policy iteration with a given base policy, once the rollout data is collected, one or two neural networks are constructed: A {\em policy network} that approximates the rollout policy, and (in the case of rollout with truncation) a {\em value network} that constructs a cost function approximation for that rollout policy (the essentially synonymous terms ``actor network" and ``critic network" are also common in the literature). We consider two methods:
\begin{enumerate}[leftmargin=*]
\item {\em Approximate rollout and PI with truncation}, where each generated policy as well as its cost function are approximated by a policy and a value network, respectively. The cost function approximation of the current policy is used to truncate the rollout trajectories that are used to train the next policy.
\item {\em Approximate rollout and PI with no truncation}, where each generated policy is approximated using a policy network, but the rollout trajectories are continued up to a large maximum number of stages (enough to make the cost of the remaining stages insignificant due to discounting) or upon reaching a termination state. Here only a policy network is used; a value network is unnecessary since there is no rollout truncation. 
\end{enumerate}
Note that as in all approximate PI schemes, the sampling of feature states used for training is subject to exploration concerns. In particular, for each policy approximation, it is important to include in the sample set $\{y^s\mid s=1,\ldots,q\}$, a subset of feature states that are ``favored" by the rollout trajectories; e.g., start from some initial subset of feature states $y^s$ and selectively add to this subset feature states that are encountered along the rollout trajectories. This is a challenging issue, which must be approached with care; see \cite{LaP03, DiL08}. 


\vspace{-0.1in}
\section{Policy Iteration With a Partitioned Architecture}
We will now discuss our partitioned architecture. It is based on a partition of the set $Y$ of feature states into disjoint sets $Y_1,\ldots,Y_N$, so that $Y=\cup_{\nu=1}^N Y_{\nu}$. We train in parallel a separate (local) policy network and a (local) value network (in the case of a scheme with truncation) using feature state data from each of the sets $Y_1,\ldots,Y_N$. Thus at each policy iteration, we use $N$ policy networks and as many as $N$ additional value networks (in some specially structured problems, including the pipeline repair problem, some of the value networks may not be needed, as we will discuss later).
 \begin{figure}[h]
    \centering
    \includegraphics[width=1.0\linewidth, height=6cm]{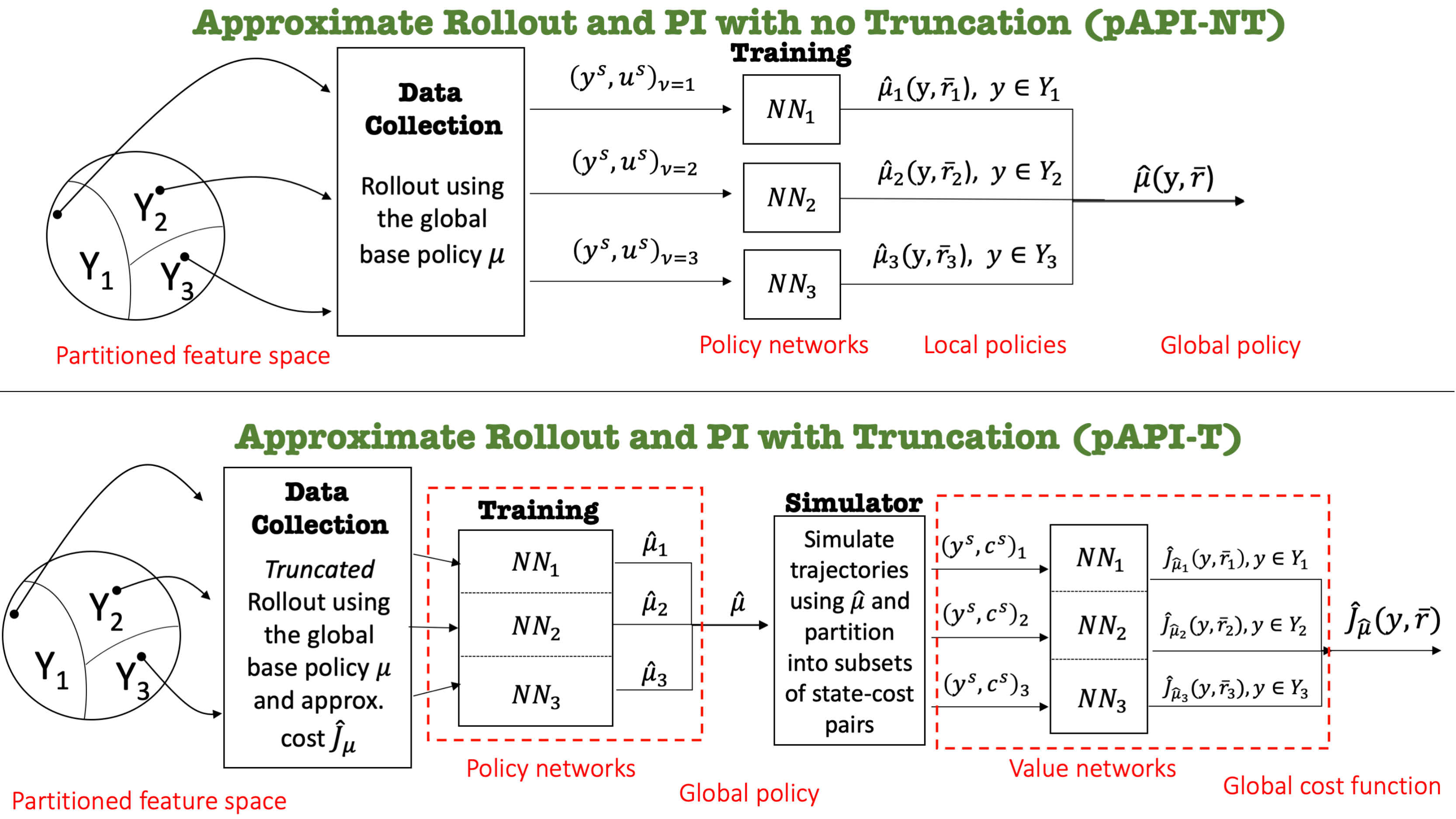}
    \caption{\vspace{-0.01in}Partitioned architecture for rollout and approximate PI. \SB{For truncated rollout, we may employ terminal cost approximation using a value network.
    \vspace{-0.04in}}}
    \label{fig:approx_pi}
\end{figure}\
The $N$ local policy networks, each defined over a subset $Y_{\nu}$, are combined into a global policy, defined over the entire feature space $Y$. For schemes with truncation, each local value network is trained using the global policy, but with starting feature states from the corresponding subset $Y_{\nu}$ of the partition; see Fig. \ref{fig:approx_pi}.

This partitioned architecture is well-suited for distributed computation, with the local networks sharing and updating piecemeal the current global policy and (in the case of truncation) the current global terminal cost approximation. Moreover, we speculate that our methodology requires smaller training sets, which cover more evenly the feature space, thereby addressing in part the issue of adequate feature space exploration.
While it is hard to quantify this potential advantage, our computational experience indicates that it is substantial. Regarding the distributed training of our partitioned architecture, we may conceptually assign one virtual processor for each set of the feature space partition (of course multiple virtual processors can coexist within the same physical processor). The empirical work described in the following sections provides, among others, guidelines for future research directions on the performance and implementations of synchronous and asynchronous distributed PI with partitioned architectures (in the spirit of \cite{BeY10}), particularly in the context of POMDP as well as multi-agent robotics.

\vspace{-0.04in}
\section{Sequential Repair and the Pipeline Problem}
Various elements of the approximate PI methodology just described have been applied to a pipeline repair problem. We will describe the implementation, computational results, and comparisons with other methodologies in detail. The problem involves autonomous repair of a number of pipeline locations where damage may have occurred, and the objective is to find a policy to repair the pipeline with minimum cost, based on available information. Our results show that 1) approximate PI can be successfully applied in the context of POMDP, 2) a partitioned architecture results in substantial computation time savings, thus allowing applicability of our methodology to large state space problems, and 3) our pipeline repair model and solution methods can be extended to complex two-dimensional and multi-\SB{robot} 
contexts, where a solution can be facilitated by using partitioning.
\vspace{-0.05in}
\subsection{Pipeline Repair Problem Description}
\begin{figure}[h]
\vspace{-0.04in}
    \centering
    \includegraphics[width=0.85\linewidth, height=4cm]{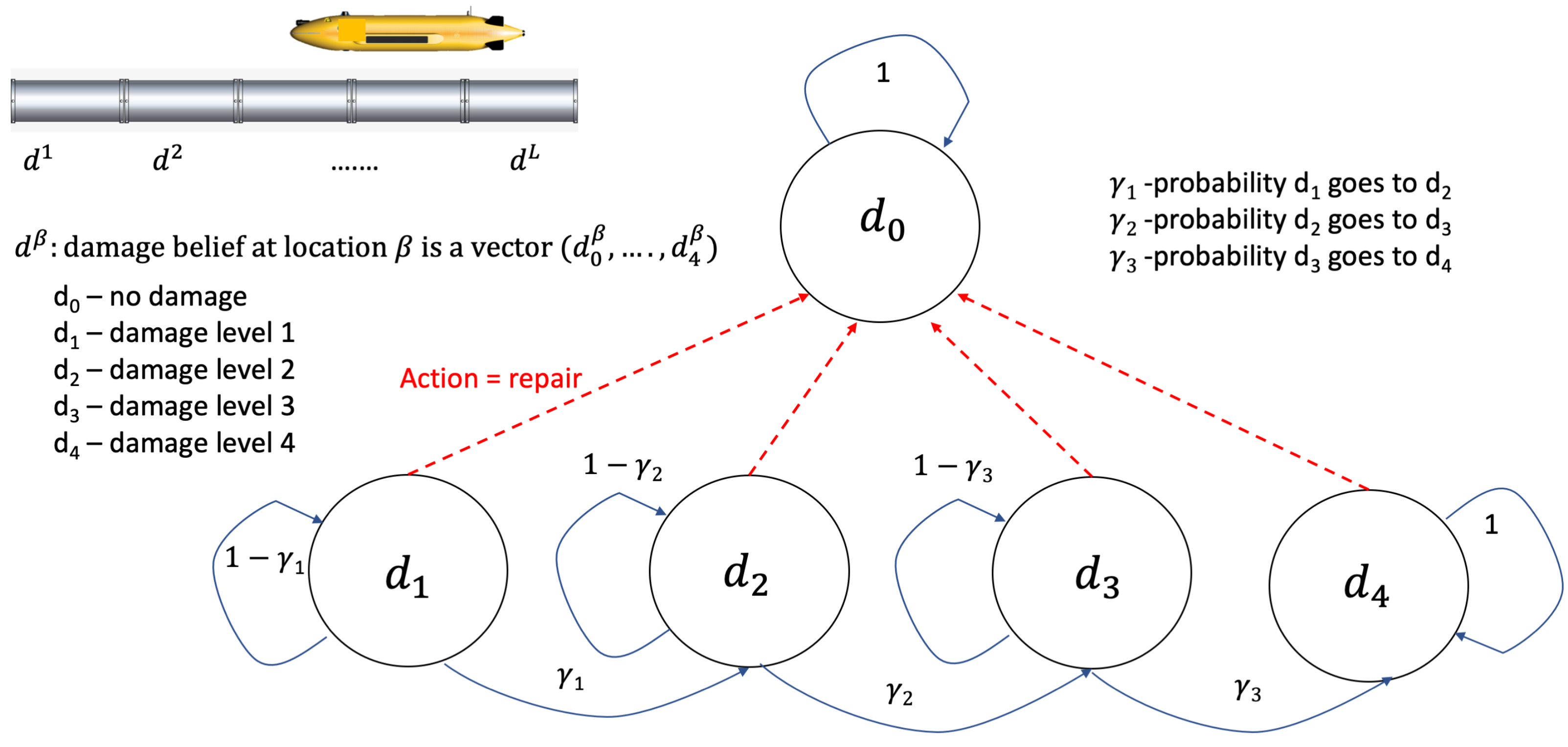}
    \caption{Markov chain for the damage level of each location of the pipeline. 
    }
    \label{fig:markov}
\end{figure}

In this problem, an autonomous robot moves along a pipeline consisting of a sequence of $L$ locations, denoted $1,\ldots,L$. Each location of the pipeline can be in one of $\zeta+1$ progressively worse damage levels indicated by $d_0,d_1,\ldots,d_\zeta$, where $d_0$ indicates a no damage condition. The damage level of each location changes stochastically over time according to a known Markov chain with the $\zeta+1$ states $0,1,\ldots,\zeta$ (see 
Fig.~\ref{fig:markov}). As the figure indicates, we assume that a location that is not damaged (state $d_0$), cannot become damaged. However, for a damaged location, the level of damage can stochastically become worse. We assume that the robot has a sensing radius of one location within which it can verify the damage level of the location. Thus with each visit to a location, an observation is obtained, namely, the exact damage level of the location. 

The robot knows its current location, and maintains for each location $\b$, a belief state of damages $d^\b=(d_0^\b,\ldots,d_\zeta^\b)$ consisting of the conditional probabilities $d_0^\b,\ldots,d_\zeta^\b$ of the damage level, given the current observation history. 
The initial belief state could be obtained from information gathered \emph{a priori}, for example, via noisy images of the pipeline. Upon reaching a location $\b$, the robot determines its damage level, and decides upon one of three actions: Stay in $\b$ for one time period and repair the damage, bringing it to level $d_0$, or move to one of the two adjacent locations $\b-1$ or $\b+1$ without repairing the damage (if $\b=1$ or $\b=L$, only two of these actions are available). There is a known cost per unit time for each location, depending on its damage level, and the objective is to minimize the discounted sum of costs of all the locations of the pipeline over an infinite horizon. This is a POMDP with $L\cdot (\zeta+1)^L$ states and three actions per state. 

It is straightforward to implement the belief state estimator, given the initial belief distributions of the damage levels of all the locations. It is also straightforward to program a simulator of the type illustrated in Fig. \ref{belief_simulator}, and to generate rollout trajectories with a given base policy starting from a given feature state.
\vspace{-0.05in}
\subsection{Partitioned Architecture for the Pipeline Repair Problem}
\begin{figure}[h]
    \centering
    \includegraphics[width=1.0\linewidth, height=4.5cm
    ]
    {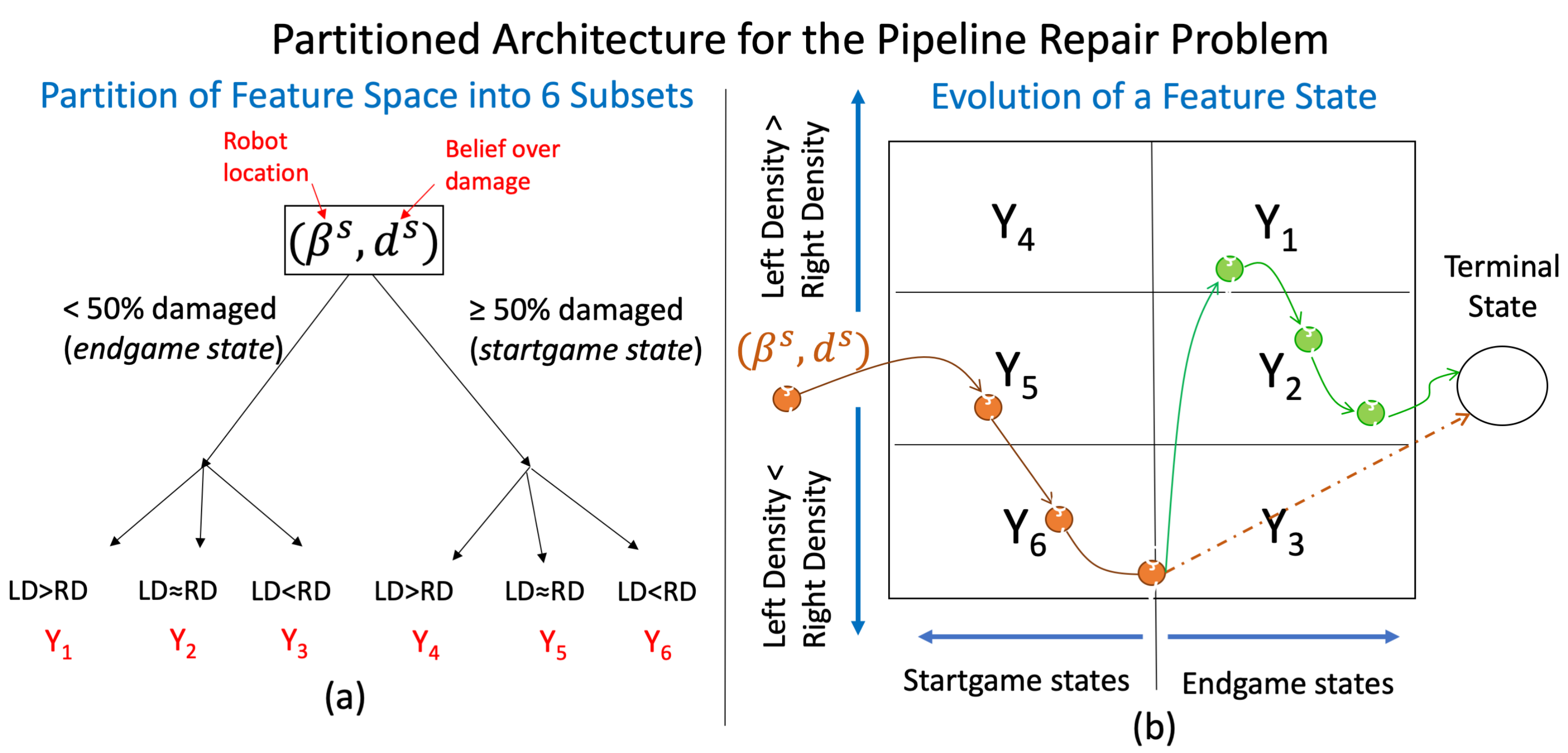}
    \caption{\vspace{-0.01in}Partition of the feature space into subsets for the pipeline problem. 
    }
    \label{fig:pattern}
    \vspace{-0.05in}
\end{figure}

For our partitioning scheme, we use two characteristics of the belief vector. These are \SB{(see Fig.~\ref{fig:pattern}a)}:\\
\noindent \textbf{Percentage of disrepair:} \SB{If less than one half of the pipeline locations are believed to be in some damaged level ($d_1, d_2, d_3, d_4$)}, we say that we are in an \emph{endgame} state, and otherwise we say that we are in a \emph{startgame} state. 
\noindent\textbf{Density of possible damage:} We distinguish three cases, based on two variables, $LD$ and $RD$ (Left and Right Damage), which will be defined shortly. The three cases are a) damage to the left is greater than to the right: $\frac{LD}{LD+RD}>0.7$, b) damage to the left is approximately equal to damage to the right: $0.3\le \frac{LD}{LD+RD}\le 0.7$, and c) damage to the right is greater than to the left: $\frac{LD}{LD+RD}<0.3$. 

Based on the six possible pairs of the characteristics above, the belief space is partitioned into six corresponding subsets, and the index of the current subset is added to the belief state $b$ to form the feature state $y$ (hence $y$ subsumes $b$ and its evolution can be sequentially simulated, in conjunction with $b$, consistent with our assumptions). Since we assume that a repaired location cannot fall into disrepair, endgame states cannot lead to startgame states. This allows a sequential training strategy whereby endgame cost and policy approximations can be used for training startgame policy and value networks.
\vspace{-0.05in}
\subsection{Algorithmic Implementation}
\label{sec:pipelineAlg}
\begin{figure}[h]
    
    \centering
    \includegraphics[width=1.0\linewidth, height=5.5cm]{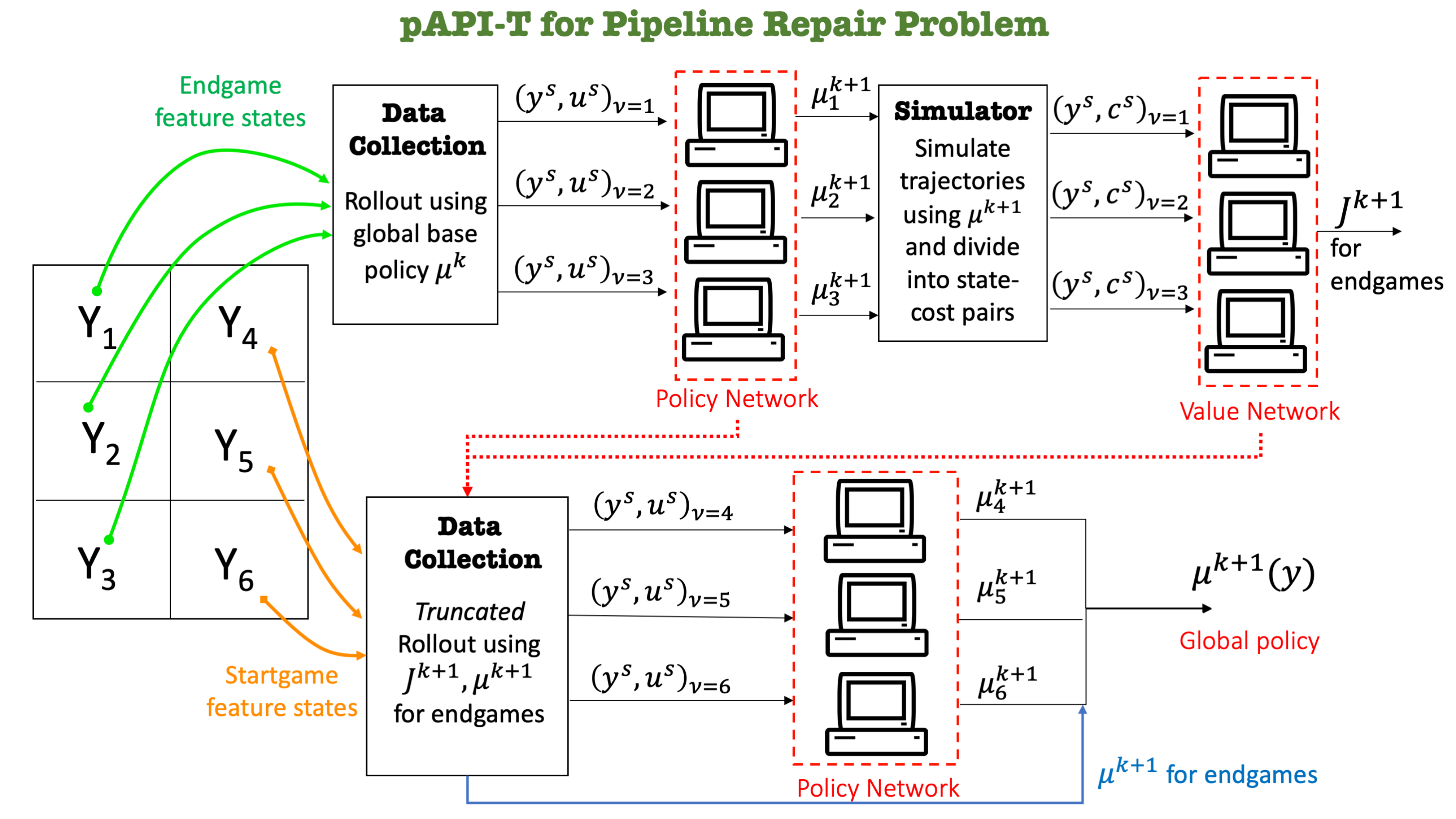}
    \caption{Algorithm for applying pAPI-T to our pipeline repair problem. 
    }
    \label{fig:pattern_pipeline}
    \vspace{-0.06in}
\end{figure}

We employ two algorithms that are based on the partitioning scheme just described: Partitioned Approximate Policy Iteration with No Truncation (pAPI-NT) and Partitioned Approximate Policy Iteration with Truncation (pAPI-T) (see Fig.~\ref{fig:pattern_pipeline}). These algorithms
 are well-suited for distributed computation, with a separate processor being in charge of the simulation and training within its own subset of the partition, and communicating its local policy and cost function to the other processors for use in their truncated rollout computations. Also, the approximate cost functions for the endgame subsets are shared as terminal cost approximations with other processors that are learning startgame policies. We find that these schemes save computation time while performing almost as well as a nonpartitioned architecture in terms of minimizing cost (see the next section).
\vspace{-0.1in}
\vspace{-0.04in}

\subsection{Experimental Results}
We implemented our proposed methodology on various problems of different sizes: 1) A linear pipeline problem of 20 locations and $10^{15}$ states, 2) a two-dimensional grid pipeline of size 6$\times$6 and $10^{26}$ states, and 3) a two-robot linear pipeline of $10^{16}$ states. We describe our experimentation for the linear pipeline first.

For our experiments we have made the following choices. We used neural networks for both policy and value approximations. 
The policy networks used 2 hidden layers of size 256 and 64 ReLU units respectively, while the value networks used 3 hidden layers of 256, 128, and 64 ReLU units.
The output layer of the policy network used a softmax layer over 3 actions (repair, go left, go right) in a linear single-robot pipeline (5 actions for a single robot 2D grid, 9 actions for two-robot linear pipeline).
We used the RMSprop optimizer to minimize the L2 loss with a learning rate of 0.001.

To obtain the rollout policy, we used one-step lookahead and 10 Monte-Carlo simulated trajectories with the base policy from each leaf of the lookahead tree. We have not tried multistep lookahead in order to keep the size of the lookahead tree manageable. 
We used a greedy policy as the starting base policy, which repairs the current location if it is damaged, moves left if out of all damaged locations, the nearest one is on the left, or moves right otherwise. The density of damage to the left of the robot at location $\b$ is defined as $LD=\sum_{j=0}^{\b-1} {d^j}' c$, where $c=[0,0.1,1,10,100]$ is the vector of costs for the different damage levels. The definition of $RD$ is similar. For the 20-location linear pipeline cases, 200,000 training samples were used for each subset of the feature space partition (and $1.2\times 10^6$ samples for training using the methods that do not use partitioning). 
For the 6$\times$6 grid and two-robot cases, 500,000 training samples were used for each subset of the partition. Training samples were obtained by: 1) random generation of feature states from each partition set (Fig.~\ref{fig:pattern}(b)), and
2) augmenting the pool of samples with representative states over a memory buffer where we randomly retained \SB{around 10\%} of states generated while evaluating the previous policies as described in \cite{Ber19}, Ch.\ 5. 
 
\begin{figure}[h]
    \centering
    \includegraphics[width=0.8\linewidth, height=4.5cm]{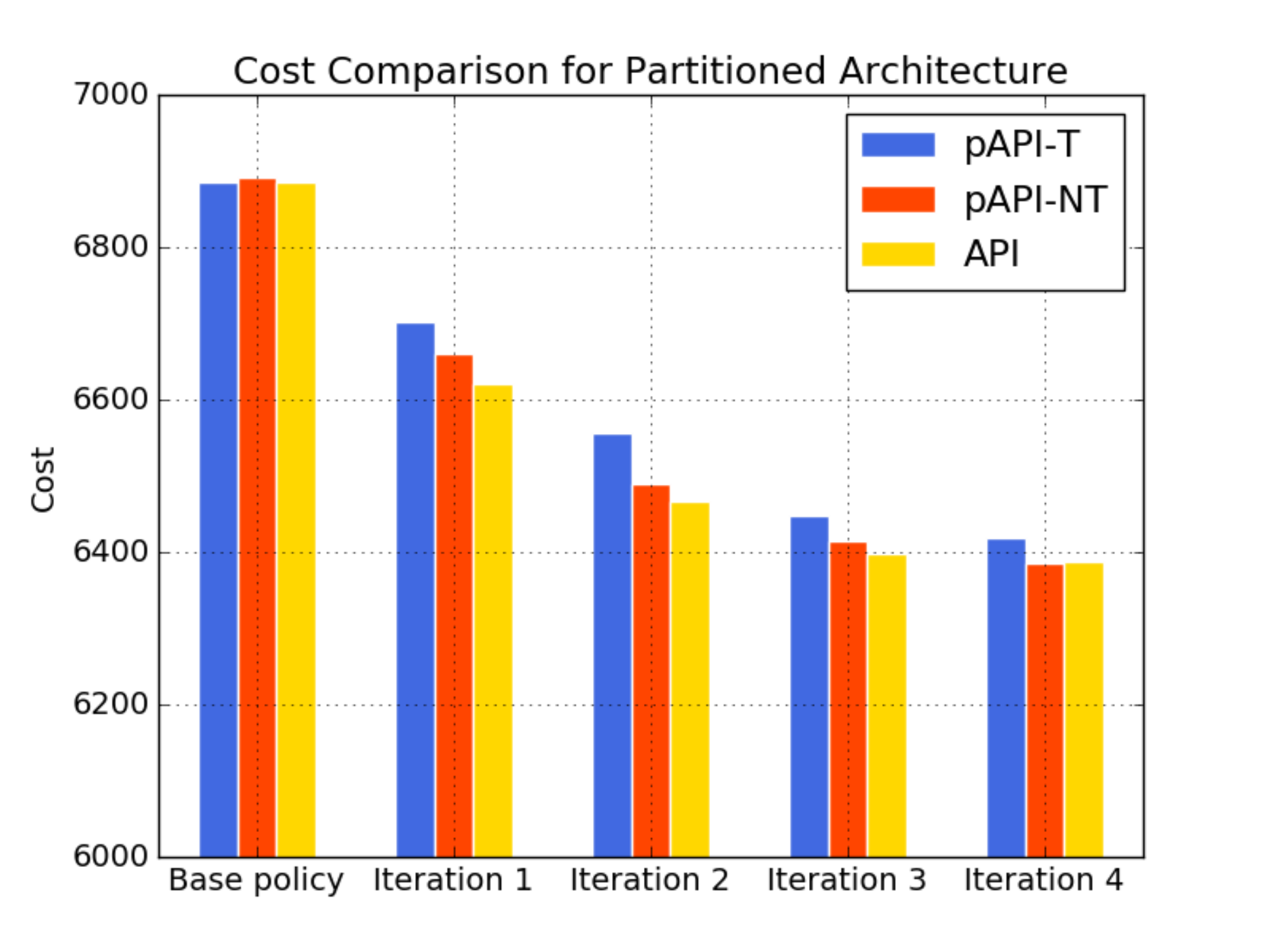}
    \caption{Performance comparison of API, where pAPI-T and pAPI-NT use partitioning, with and without truncation respectively.
    }
    \label{fig:architectureComparisons}
\end{figure}

We have generally observed that the partitioned architecture results in significant improvement in running time thanks to parallelization, at the expense of a modest degradation in performance. We expect that this improvement will become more significant as we address larger problems that require a richer feature space partition. 

\begin{figure}[h]
    \centering
    \includegraphics[width=1\linewidth, height=5cm]{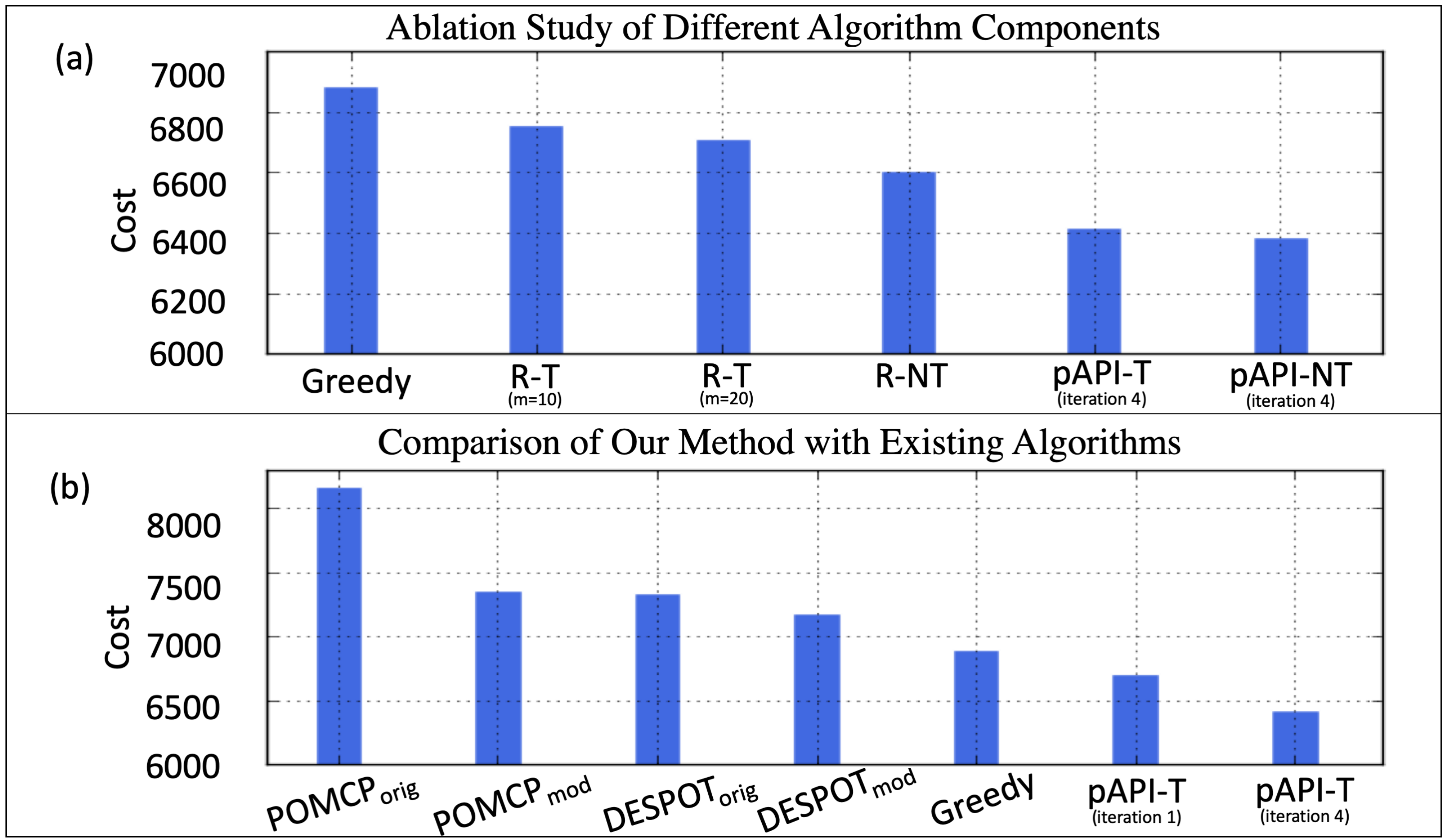}
    \caption{(a) Ablation study demonstrating performance contributions of the different components of our methodology. (b) Comparison of our methods to DESPOT and POMCP. 
    }
    \label{fig:Ablation}
\end{figure}

We will now present an ablation study of our proposed method that demonstrates the performance of the major pieces of our algorithm. In particular, we compare: 1) \SB{pAPI-T and pAPI-NT}, which are the full versions of our algorithm with and without truncation respectively, 2) \SB{R-T and R-NT}, which are rollout algorithms with and without truncation, which are the basis for data collection and training for the improved policies used in \SB{pAPI-T and pAPI-NT}, and finally 3) the greedy policy, which is used as the starting policy for rollout and for PI. The plot in Fig.~\ref{fig:Ablation}(a) shows how each component of our method adds to improved performance. Notice that pAPI-T uses a terminal cost approximation and results in computational savings over pAPI-NT, but incurs a slight approximation penalty.

\subsubsection{Comparison to Existing Methods}
\label{sec:resultsSOTA}
We have compared our method with two published POMDP methods: DESPOT and POMCP. Our simulations were conducted using the code for these methods, which is 
freely available at~\cite{despot}. These methods do not use rollout with a base policy, but instead rely on long lookahead supplemented by Monte Carlo tree search and/or heuristic pruning of the lookahead tree. Results are shown in Fig. \ref{fig:Ablation}(b) for an evaluation set of 1000 random startgame states. We tried many DESPOT parameter combinations, including $K=100,200,300$, $D=60,90,100$, and $\lambda=0.1,0.3$. Fig.~\ref{fig:Ablation}(b) shows the best results that we obtained with DESPOT using $D=90$, $K=200$, 
\SB{$\lambda=0.1$}
, and with POMCP using the default parameters. 
We used a closed form of belief update equation based on the Markov chain of Fig.~\ref{fig:markov}, by modifying the POMCP and DESPOT code to use a single particle to represent the belief with a weight of 1.

Our results show the cost of two implementations of POMCP and DESPOT; the first implementation denoted $\text{DESPOT}_{\textbf{orig}}$ is ``straight out of the box,'' using the code provided in~\cite{despot}. We obtained the other implementation, $\text{DESPOT}_{\textbf{mod}}$, by modifying the first to disallow ``repair'' actions at already repaired locations. We note that without this modification the DESPOT and POMCP algorithms will sometimes choose a repair action on an already repaired location, leading to higher costs. Our results indicate that pAPI-NT and pAPI-T outperform DESPOT and POMCP (original and modified versions) for our pipeline repair problem. The greedy policy, somewhat unexpectedly, also performs better than DESPOT and POMCP. The reason may be that the pipeline repair problem has a long effective planning horizon. Intuitively for such problems, DESPOT and POMCP are handicapped by their reliance on a tradeoff between lookahead length and the sparsity of the lookahead tree. Since both DESPOT and POMCP estimate Q-values approximately, with a larger and sparser lookahead, they can perform worse than the greedy base policy. By contrast, our rollout policy has the cost improvement property described in Prop.\ 1, and improves over the base policy (approximately, within a bound). 

Indeed, we have observed that for states requiring a relatively long horizon to repair the pipeline, POMCP and DESPOT often make poor decisions. However, we could find no reliable method to detect whether a particular state would lead to relatively good or bad performance for DESPOT of POMCP, except in hindsight. Nonetheless for investigation purposes, even after hand-selecting for evaluation a favorable set of states, we found that DESPOT or POMCP were still outperformed by our method. In fact, based on examination of sample generated trajectories, we have concluded that the advantage that our method holds is principally due to the long horizon exploration that is characteristic of the use of rollout with terminal cost approximation.

\subsubsection{Extensions of the Pipeline Repair Problem}
\label{sec:resultsExtensions}
We present in Table I results for two extensions of our problem to demonstrate the generality of our methodology. Specifically we show results for: 1) a two-dimensional grid pipeline and 2) a two-robot implementation of our methodology. Generally, our results show that both pAPI-T and pAPI-NT can be applied to problems with larger state space and multiple robots thanks to the use of partitioning and the attendant computational savings.


\begin{table}
    \centering
    \caption{Extensions to Pipeline Repair Problem}
    \label{Extensions}
    \begin{tabular}{|p{40mm}|p{9mm}|p{9mm}|p{9mm}|}
    \hline
    \textbf{Extension} & \textbf{Greedy} & \textbf{Iter 1} & \textbf{Iter 2}\\
        \hline
         pAPI-NT on 5x4 grid &5476.79  & 5255.64 & 4952.26\\ \hline
         pAPI-NT on 6x6 grid &18799.5 & 18037.1 & 17341.9 \\ \hline
         pAPI-NT 2-robot, linear ($L=20$) & 4142.84 & 3321.95 & 3133.08\\
         \hline
    \end{tabular}
\end{table}

\vspace{-0.1in}
\section{Concluding Remarks}
This paper develops rollout algorithms and PI methods that are well-suited to deal with the challenges of POMDP, including large state spaces, incomplete information, and long planning horizons. Thus these methods are of high relevance to robotic tasks in uncertain environments such as search and rescue. While several of the components of our methodology have been suggested for perfect state information problems, they have not been combined and adapted to POMDP. Our methods are based on partitioning the feature space and training local policies that are specialized, easier to train, can be combined to provide a global policy over the entire space, and are amenable to a highly distributed implementation.

We have applied our methods in simulation to a class of sequential repair problems where a robot inspects and repairs a linear pipeline with potentially several rupture sites under partial information about the state of the pipeline (acquired from \emph{a priori} obtained knowledge and in situ observations).
Our method's partitioning of the state space has important implications for robotics problems -- namely, it allows for massive parallelization over several potentially independent processors (or robots). Importantly, this framework lends itself to distributed learning about the environment where different partitions can correspond to states in spatially different areas of the world, thus suggesting a new basis from which to solve future multi-robot POMDP problems. 

We finally note that there are several possible extensions to our sequential repair problem, such as for example allowing for stochastic repair times, two-dimensional problems with obstacles, and multiple robots.






\vspace{0.1in}
\noindent \small{ {\textbf{Acknowledgement:}} 
We acknowledge Calvin Norman and Siva Kailas for their help with numerical studies.}


\bibliographystyle{IEEEtran}
\bibliography{IEEEabrv,references_sushmita}

\end{document}